# Machine Learning-based Layer-wise Detection of Overheating Anomaly in LPBF using Photodiode Data

Nazmul Hasan[a], Apurba Kumar Saha[a], Andrew Wessman[b], Mohammed Shafae[a,*]

*[a]Department of Systems and Industrial Engineering, The University of Arizona, Tucson, AZ 85721, USA*
*[b]Department of Materials Science and Engineering, The University of Arizona, Tucson, AZ 85721, USA*

* Corresponding author. *E-mail address:* shafae1@arizona.edu

---

**Abstract**

Overheating anomaly detection is essential for the quality and reliability of parts produced by laser powder bed fusion (LPBF) additive manufacturing (AM). In this research, we focus on the detection of overheating anomalies which can lead to various defects in the part including geometric distortion, and poor surface roughness, among others, using photodiode sensor data. Photodiode sensors can collect high-frequency data from the melt pool, reflecting the process dynamics and thermal history. Hence, the proposed method offers a machine learning (ML) framework to utilize photodiode sensor data for layer-wise detection of overheating anomalies. In doing so, three sets of features are extracted from the raw photodiode data: MSMM (mean, standard deviation, median, maximum), MSQ (mean, standard deviation, quartiles), and MSD (mean, standard deviation, deciles). These three datasets are used to train several ML classifiers. Cost-sensitive learning is used to handle the class imbalance between the "anomalous" layers (affected by overheating) and "nominal" layers in the benchmark dataset. To boost detection accuracy, our proposed ML framework involves utilizing the majority voting ensemble (MVE) approach. First, the top three ML classifiers are identified from an initial pool of classifiers, based on their performance in $k$-fold cross-validation. Next, final predictions are generated using majority voting from the individual predictions of the top three classifiers. We performed 100 iterations to generate statistically reliable results. This proposed method is demonstrated using a case study including an open benchmark dataset of photodiode measurements from an LPBF specimen with deliberate overheating anomalies at some layers. The results from the case study demonstrate that the MSD features yield the best performance for all classifiers, and the MVE classifier (with a mean $F_1$-score of 0.8654) surpasses the individual ML classifiers. Moreover, our machine learning methodology achieves superior results (9.66% improvement in mean $F_1$-score) in detecting layer-wise overheating anomalies, surpassing the existing methods in the literature that use the same benchmark dataset. Finally, based on our results, we provide useful insights and recommendations for future research on applying machine learning techniques for defect detection in AM.

---

## 1. Introduction

Laser Powder Bed Fusion (LPBF) is a metal additive manufacturing (AM) technique used to fabricate parts from a digital 3D model. This process involves selectively melting fine metal powders layer by layer to form the desired object. Its significance lies in the ability to produce intricately designed components with unmatched complexity, enabling rapid prototyping, high-value component production, and reducing material waste [1]. LPBF also offers excellent mechanical properties for the fabricated parts, which makes it suitable for critical industries such as aerospace, medical, and automotive [2].

However, despite its numerous advantages, the LPBF process is not without its challenges. One of the most critical issues is the occurrence of defects during the fabrication process including porosity, cracking, and geometric distortion, among others [3]. These defects can have severe implications, particularly in industries where part failure is not an option [4]. For instance, in the aerospace industry, a single defect can lead to catastrophic failures, resulting in significant financial losses and, more importantly, endangering human lives. Due to the layer-wise nature of the process, the defects are not always visible once the part production is completed. To mitigate these risks, close quality control is essential, and in-process monitoring has emerged as a promising solution. By introducing an online process monitoring system, the part quality can be monitored during the build [5].

Grasso and Colosimo (2017) [6] and Yadav et al. (2020) [7] summarized the available in-process monitoring systems for LPBF processes. In-process monitoring sensors can be divided into two systems: acoustic and optical [7]. Acoustic systems are difficult to calibrate because of the high noise factor. Hence, optical systems like cameras and photodiodes are more commonly used in LPBF systems. Photodiode sensors have shown great potential due to their high sampling rate and sensitivity to changes in the process conditions [8, 9]. Researchers have utilized photodiode data to train machine learning (ML) models to detect anomalies and primarily focused on porosity. Consequently, there is a noticeable gap in the literature when it comes to the detection of overheating using photodiode sensors. Overheating during the LPBF process can lead to various defects, including geometric distortion and poor surface quality [10, 11]. Detection of overheated layers during the LPBF process can open the possibilities for corrective measures upon detection and thus can contribute to the reduction of defective parts.

Gronle et al. (2023) [12] published a dataset that includes photodiode measurements collected during the fabrication of an LPBF specimen in which overheating was induced purposefully. Details on this dataset are reported in Sub-section 3.1. This dataset provides a unique opportunity to develop and test new machine learning techniques for the detection of overheating in the LPBF process. In this research, we aim to utilize this benchmark dataset to develop machine learning techniques for the detection of overheating anomalies in the LPBF process. By doing so, we hope to contribute to the ongoing efforts to improve the quality and reliability of LPBF-fabricated parts, ultimately advancing the state of the art in metal additive manufacturing.

The rest of the paper is structured as follows. In Section 2, we review the previous studies on the use of photodiode signals and machine learning for in-process monitoring of the LPBF process. In Section 3, we introduce the benchmark dataset, the classification problem, and our machine learning approach. We present and analyze the results of our experiments in Section 4. We conclude with a summary and some future directions in Section 5.

## 2. Literature Review

Researchers have explored the application of machine learning techniques to detect anomalies and defects in LPBF-printed parts by utilizing photodiode signals. Okaro et al. (2019) [13] employed a semi-supervised machine learning approach to detect defects in tensile test bars by using photodiode data. Mao et al. (2023) [9] proposed a method based on machine learning algorithms. Their method combines the photodiode signal and melt pool temperature to detect anomalies in the melt pool temperature field. Their work is inspired by the insights of Feng et al. (2022) [14] who demonstrated that analysis of the high-resolution melt pool temperature field can be leveraged to detect defects like surface porosity and spatter. Inspired by these findings, Mao et al. (2023) aimed to establish a correlation between the photodiode signal and melt pool temperature to allow defect detection using cost-effective photodiode sensors. Taherkhani et al. [15] utilized photodiode measurements to exploit variations in light intensity emitted from the melt pool, enabling the detection of the lack of fusion porosities identified through micro-computed tomography scanning. The authors conducted experiments with intentionally seeded defects and stochastic distributions of pores to evaluate the effectiveness of their approach. The results demonstrate the detectability of voids and the correlation between sensor predictions and randomized defects under different sets of process parameters. Tao et al. (2023) [16] conducted a parallel study, exploring the predictive capabilities of porosity by integrating photodiode signals with X-ray computed tomography (X-CT). They recorded photodiode signals from a coaxial melt pool monitoring system and utilized X-CT to collect data on the size, shape, location, and number of pores. Next, they used layer-wise segmented photodiode signals to train machine learning models to predict the presence and location of porosities with reasonable accuracy. These research studies collectively underscore the potential of machine learning techniques in identifying anomalies and defects through the utilization of photodiode signals in the LPBF process. However, a significant gap remains in employing machine learning to detect overheating anomalies via photodiode signals.

Gronle et al. (2023) [12] proposed a benchmark dataset for the detection of overheating anomalies. As a sample analysis, the authors employed a layer-wise statistical monitoring approach using univariate control charts to flag overheating anomalies. However, because of the inherent limitations of the univariate control chart, the authors declared it as an open problem and made the dataset publicly available to support the development of superior methodologies. Chen et al. [17]

introduced a complex hyperdimensional computing (HDC) methodology for overheating anomaly detection, utilizing the benchmark dataset provided by Gronle et al. (2023) [12]. Their HDC framework uses temporal sampling of photodiode signal data and yields a moderate $F_1$-score (0.745±0.214), leaving room for further improvement. In response, our research presents an ensemble machine learning framework, designed to enhance the detection of overheating anomalies utilizing the same benchmark dataset. Additionally, the methodology of Chen et al. [17] has a few other limitations including the use of under sampling technique to deal with the class imbalance allowing loss of valuable information from the majority class, which may lead to less accurate models. To address this issue, we aim to use cost-sensitive learning which is a superior alternative to sampling methods for imbalanced learning [18]. We believe that our proposed method provides a robust pathway for detecting overheating in LPBF using photodiode measurements.

## 3. Methodology

In this section, we first describe the benchmark dataset that is utilized to develop machine learning classifiers for overheating anomaly detection. Subsequently, we explain our approach in detail encompassing discussions on feature extraction, managing imbalanced data, and ensemble learning through majority voting.

### 3.1. Dataset

The benchmark dataset contains signals acquired via spatially integrated indium gallium arsenide (InGaAs) photodiode sensor during the manufacturing of an aluminum alloy (AlSi10Mg) specimen using a multi-laser LPBF Trumpf system [12]. The specimen is a parallelepiped of size 10×10×25 mm and it was printed with fixed process parameters (laser spot diameter of 100 µm, laser power of 480 W, and a scan speed of 1500 mm/s). Fig. 1 shows a schematic view of the fabricated specimen.

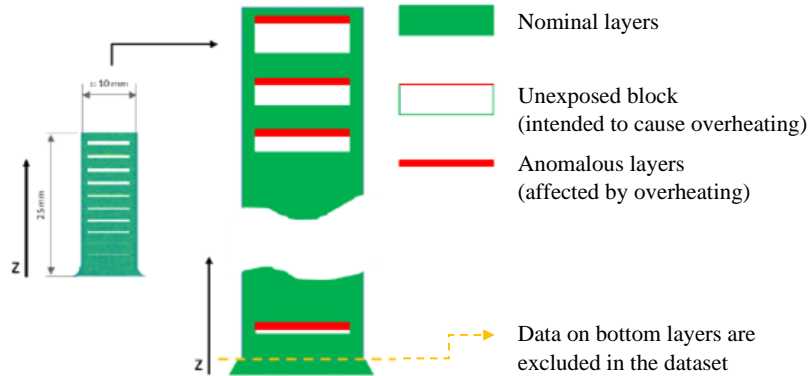

Fig. 1. Schematic view of the LPBF printed part with unexposed blocks to force overheating anomaly [12]

Overheating anomalies were deliberately induced by strategically incorporating multiple unexposed blocks into the part geometry. In Fig. 2, the spatial distribution of photodiode measurements illustrates a sample unexposed block layer, an anomalous layer affected by overheating, and a nominal bulk layer. The design of unexposed blocks ensures laser scans exclusively cover the periphery of the square cross-section, leaving the central region unexposed (Fig. 2(a)). Consequently, the subsequent fully laser-scanned layer exhibits an overhang region at the center, leading to overheating (Fig. 2(b)). This overheating effect cascades to subsequent layers. The dataset's author posits that - three layers following the unexposed blocks experience overheating, with a return to normalcy from the fourth layer onward. Fig. 2(c) visually supports this assumption. Notably, the top limit in the color bars of Fig. 2 is set at 1800 for effective visual comparison of the overheating effect across different layers.

The benchmark dataset comprises photodiode measurements for the upper 379 layers, starting above the tapered base of the specimen. The number of consecutive layers within these unexposed blocks varies from 1 to 10 along the z-direction, totaling 55 layers in the entire specimen. Post-process X-CT reveals an absence of overhang regions after the block with one unexposed layer. This observation can be attributed to the distinct remelting effect inherent in the LPBF process. Consequently, no overheating anomalies were considered in layers after the first unexposed block. Thus, the dataset includes 27 (=(10−1)×3) anomalous layers, while the remaining 352 (=55+297) layers are considered nominal bulk layers.

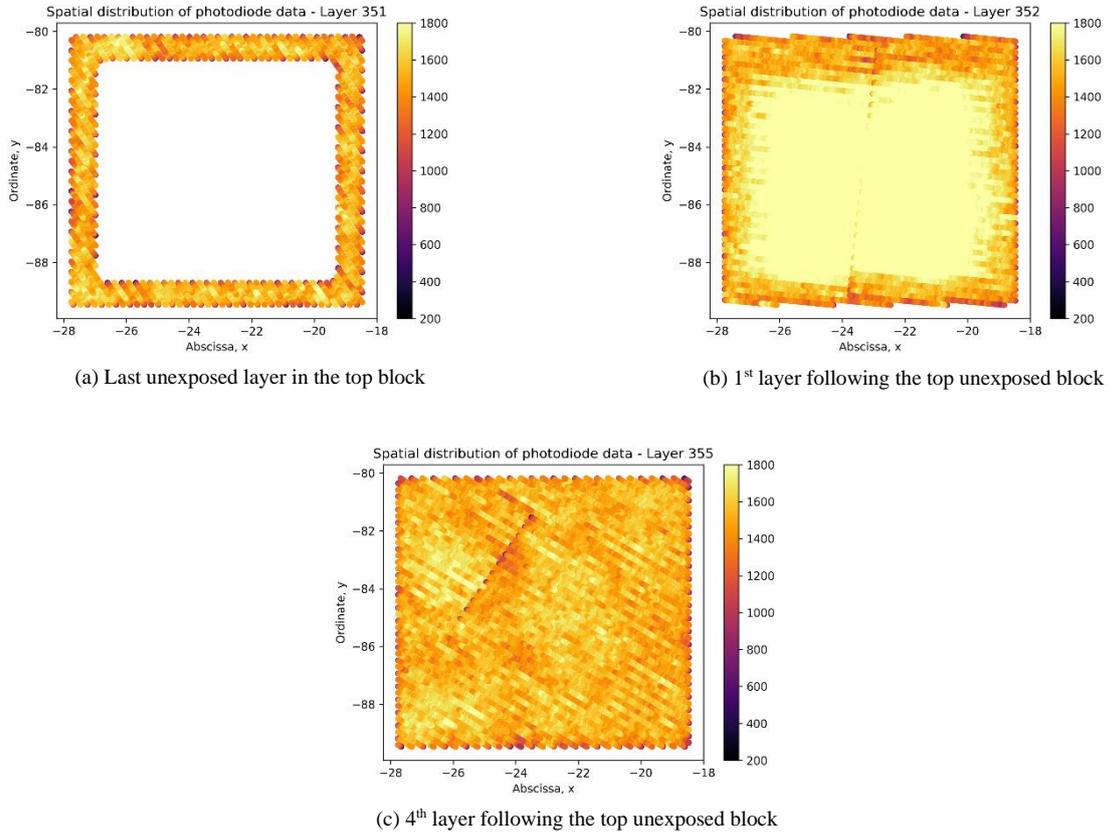

(a) Last unexposed layer in the top block

(b) 1st layer following the top unexposed block

(c) 4th layer following the top unexposed block

Fig. 2. Spatial distribution of the raw InGaAs photodiode signals across three types of layers: (a) unexposed block layer; (b) anomalous layer affected by overheating; (c) nominal bulk layer.

In their respective works, Gronle et al. (2023) [12] and Chen et al. (2023) [17] focused on 297 nominal bulk layers and 27 anomalous layers, excluding unexposed block layers from their analysis. However, we argue that unexposed block layers can offer insights into specimen geometry variations and should be treated as nominal layers, given their expected immunity to overheating. Consequently, our study addresses a classification problem with 352 instances under the "Nominal" class and 27 instances under the "Anomalous" class, representing a typical anomaly detection scenario with class imbalance. To address this, we employ cost-sensitive learning, which is discussed in detail in Sub-section 3.3.

### 3.2. Feature extraction

Fig. 3 displays the raw InGaAs photodiode signals corresponding to the three layers depicted in Fig. 2. Notably, the length of the InGaAs photodiode signal varies across layers, contingent on the scanned area. The signal array length ranges from 30,339 to 31,135 for bulk layers, 9,560 to 9,726 for unexposed block layers, and 30,483 to 31,102 for anomalous layers. Fig. 4 depicts a visualization showcasing the nominal and anomalous layers within the feature space defined by the mean and standard deviation of the photodiode measurements. The limited differentiation between the two classes in this two-dimensional feature space emphasizes the necessity for incorporating additional features and employing advanced tools such as machine learning techniques, surpassing the capabilities of univariate control charts. In this research, we opted to extract three progressively refined sets of characteristic features from the signal measurements:

(1) mean, standard deviation, median, maximum (MSMM)
(2) mean, standard deviation, and quartiles (MSQ)
(3) mean, standard deviation, and deciles (MSD)

The MSMM feature set comprises four global features, while the MSQ feature set is a superset of MSMM, including two additional local features. The MSD feature set is even more granular, containing additional local features. These three feature datasets are employed to train machine learning classifiers. The utilization of these progressively refined feature sets aims to explore whether more granular features contribute to more effective classification or potentially suffer from overfitting.

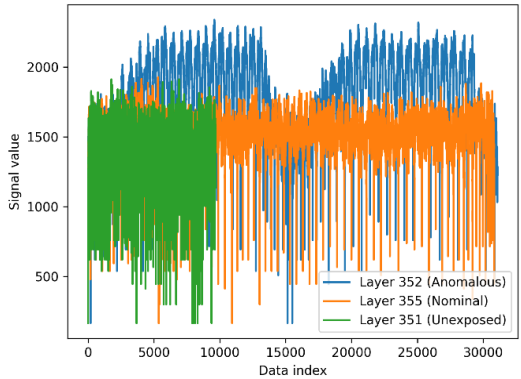

Fig. 3. Raw InGaAs photodiode signals across different layers

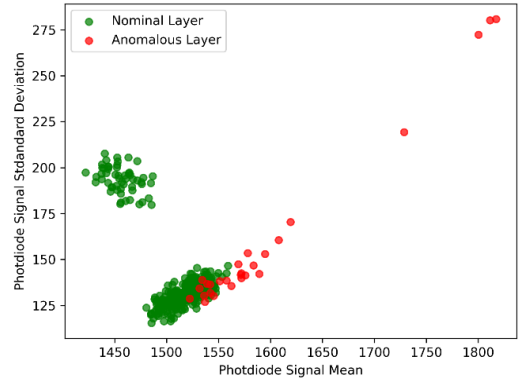

Fig. 4. Visualization of the nominal and anomalous layers in photodiode signal mean and standard deviation feature space.

### 3.3. Handling class imbalance

As mentioned in Sub-section 3.1, our classification problem has 352 instances under the "Nominal" class and 27 instances under the "Anomalous" class. Methods for handling class imbalance can be grouped into three categories [19]: data-level methods, algorithm-level methods, and hybrid methods. Data-level methods try to handle the class imbalance by either increasing the number of samples from the minority class (over-sampling) or decreasing the number of samples from the majority class (under-sampling). These methods are simple and easy to use, but they have some drawbacks. Over-sampling may not effectively boost existing rare cases, and adds no new information, while under-sampling can harm performance by removing vital information from the majority class [20]. On the other hand, algorithm-level methods try to modify the learning algorithm to make it more sensitive to the minority class. They can either assign different weights or costs to the classes (cost-sensitive learning) or adapt the decision threshold or the learning parameters of the algorithm to reduce bias towards the majority class [19]. The hybrid methods try to combine data-level and algorithm-level techniques.

In this study, we chose to implement cost-sensitive learning, an algorithm-level approach. This decision stems from empirical findings in imbalanced learning domains, where cost-sensitive learning has demonstrated superior performance compared to sampling methods [18]. Furthermore, our choice is motivated by the intention to prevent information loss through undersampling and the introduction of noise in the training dataset via oversampling.

To implement cost-sensitive learning, we aim to use machine learning classifiers capable of incorporating weighted loss functions, such as logistic regression (LR), support vector classifier (SVC), random forest (RF), and decision tree (DT). To ensure objectivity, we employ "balanced" weights which refers to the assignment of weights to classes in inverse proportion to their frequencies. It is worth noting that assigning higher weights for a class is equivalent to assigning higher costs for the misclassification of that class. A detailed summary of the parameters of selected machine learning classifiers is reported in Table 1.

Table 1. Parameter setting for machine learning classifiers

| ML Classifier | Parameters |
| --- | --- |
| Random Forest (RF) | Class weight: Balanced, Number of estimators: 200, Maximum depth: 20, Minimum number of samples to split a node: 2, Minimum number of samples to be at a leaf node: 1 |
| Decision Tree (DT) | Class weight: Balanced, Maximum depth: 20, Minimum number of samples to split a node: 2, Minimum number of samples to be at a leaf node: 1 |
| Logistic Regression (LR) | Class weight: Balanced, Inverse of regularization strength: 1.0, Maximum number of iterations: 1000 |
| Support Vector Classifier (SVC) | Class weight: Balanced, Inverse of regularization strength: 1.0, Kernel function: Radial basis function, Kernel coefficient: 'scale', Enable probability estimates: True |

*3.4. Ensemble learning*

Ensemble learning is a machine learning technique that combines the predictions of multiple models to achieve better performance than any single model alone [21]. Ensemble learning can be useful for dealing with complex problems that require high accuracy, robustness, and generalization [22]. One of the most common methods of ensemble learning is majority voting ensemble (MVE) [23], which is also known as hard voting. In majority voting, each model in the ensemble casts a vote for a class label, and the class label with the most votes is chosen as the final prediction. Majority voting can be seen as a way of aggregating the opinions of different experts and reducing the effect of individual errors [24].

Algorithm 1 outlines our ensemble learning methodology, employing MVE to determine class labels. First, the dataset containing features and class labels is split into 70% training and 30% testing subsets, maintaining a balanced class ratio. Subsequently, we employ *k*-fold cross-validation exclusively on the training data, preserving class balance within each fold. Through *k*-fold cross-validation, we iteratively train multiple (>3) machine learning classifiers, ensuring that their performance is evaluated on a validation set for each fold. The average performance metric is then computed across all folds. Note that the selected pool of machine learning classifiers must be capable of cost-sensitive learning [18]. The top three classifiers are identified based on this metric. Following this, the top three classifiers are trained on the entire training data. Next, their predictions on the testing data are aggregated using the MVE approach to yield the final class labels for the testing data. Thus, our methodology combines rigorous cross-validation, classifier selection, and ensemble techniques to enhance the robustness and accuracy of the final predictions.

**Algorithm 1.** Majority Voting Ensemble (MVE)-based framework

**Input:** Dataset (containing features and class labels); Multiple (>3) machine learning classifiers (capable of cost-sensitive learning); Performance metric (for assessing the classifiers).

**Output:** Class labels (for testing data).

**Framework steps:**

1. Split the entire dataset, preserving a balanced ratio of the classes, into 70% training and 30% testing data subsets.
2. Perform *k*-fold cross-validation exclusively on the training data:
   a. Divide the training data into *k* folds, preserving a balanced ratio of the classes.
   b. Iterate through each fold as the validation set, while the remaining folds constitute the training set.
   c. Train the cost-sensitive learning classifiers using the training set and evaluate their performance on the validation set.
   d. Repeat steps b-c for all folds and compute the average performance metric.
3. Identify the top three classifiers based on the value of the average performance metric.
4. Generate predictions from the top three classifiers:
   a. Train each of the top three classifiers on the entire training data.
   b. Use these classifiers to predict the classes for the testing data.
5. Perform majority voting on the predictions generated by the top three classifiers to yield the final class labels.

## 4. Results and discussion

In this section, we implement the ensemble learning methodology, as outlined in Algorithm 1, to address the classification problem introduced in Sub-section 3.1, aiming to establish a robust machine learning technique for detecting overheating anomalies in the LPBF process. As reported in Algorithm 1, the ensemble learning methodology has 3 inputs: dataset (containing features and class labels); multiple (>3) machine learning classifiers (capable of incorporating weighted loss functions); and performance metric (for assessing the classifiers).

We use the three feature sets detailed in Sub-section 3.2: MSMM (mean, standard deviation, median, maximum), MSQ (mean, standard deviation, quartiles), and MSD (mean, standard deviation, deciles). The class labels are derived from the benchmark dataset introduced in Sub-section 3.1. This dataset includes 352 instances of the "Nominal" class and 27 instances of the "Anomalous" class. Consequently, the combination of the three feature sets with the corresponding class labels forms three distinct datasets for our analysis.

### 4.1. Machine learning with cost-sensitive learning

We utilize the machine learning classifiers detailed in Table 1, all of which support a weighted loss function to address the inherent class imbalance, a critical concern in our classification problem. The $F_1$-score (Eq. (3)) serves as our primary performance metric during the identification of the top three classifiers post the $k$-fold cross-validation phase. $F_1$-score is prioritized over accuracy (Eq. (4)) in the context of imbalanced data due to its consideration of both precision (Eq. (1)), and recall (Eq. (2)), effectively striking a balance between the two [19]. In contrast, accuracy lacks sensitivity to class distribution and can be deceptive when confronted with imbalanced classes.

$$\text{Precision} = \frac{\text{TP}}{\text{TP} + \text{FP}} \qquad (1)$$

$$\text{Recall} = \frac{\text{TP}}{\text{TP} + \text{FN}} \qquad (2)$$

$$F_1 = \frac{2 \times \text{Precision} \times \text{Recall}}{\text{Precision} + \text{Recall}} \qquad (3)$$

$$\text{Accuracy} = \frac{\text{TP} + \text{TN}}{\text{TP} + \text{TN} + \text{FP} + \text{FN}} \qquad (4)$$

Here, TP, TN, FP, and FN denote true positive, true negative, false positive, and false negative class labels, respectively. For $k$-fold cross-validation, we choose $k = 5$ to ensure an ample representation of instances from the minority "Anomalous" class in the validation data. To guarantee the statistical reliability of our classification results, we apply the ensemble learning methodology through 100 iterations on each of the three datasets: MSMM, MSQ, and MSD.

In Fig. 5, the ranking of the top three classifiers based on the $F_1$-score in $k$-fold cross-validation under the three datasets is illustrated. Notably, the random forest (RF) classifier consistently achieves the top rank across all three datasets over the 100 iterations, highlighting its superiority and robustness for the current classification problem. Logistic regression (LR) and decision tree (DT) classifiers alternately share the 2nd and 3rd ranks across different datasets throughout the 100 iterations. However, the support vector classifier (SVC) obtains the 3rd rank only in a few iterations under the MSD feature set, indicating that SVC may not be as suitable for detecting overheating anomalies in this context.

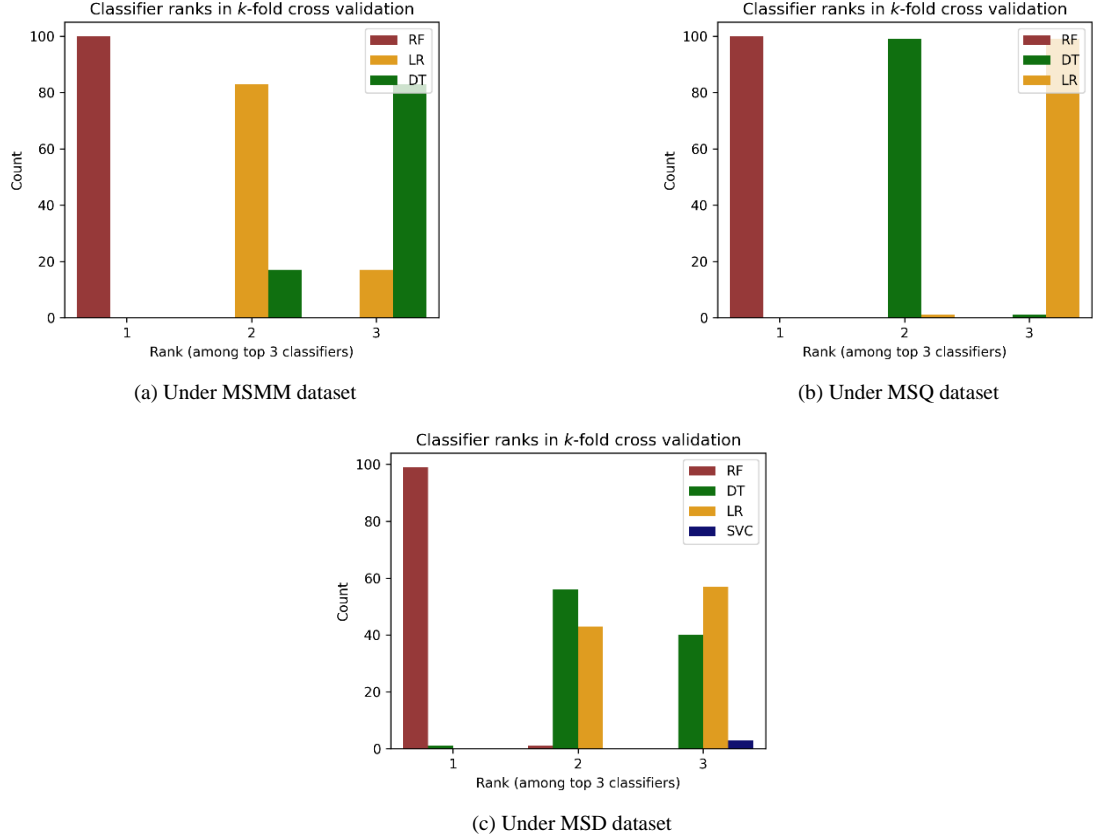

Fig. 5. Rank of the top three classifiers based on $F_1$-score in $k$-fold cross-validation under the three datasets: (a) MSMM; (b) MSQ; (c) MSD.

In Table 2, the mean and standard deviation of the $F_1$-score and accuracy metrics, obtained from 100 iterations, are showcased. The results are detailed for both MVE and individual machine learning classifiers across the three distinct datasets. Notably, the maximum mean values for the metrics within each dataset are emphasized through bold font, while the overall highest mean metric value is underlined for clarity. While the $F_1$-score remains our primary metric for assessing classification results, the inclusion of accuracy in Table 2 provides a comprehensive overview of the classification performance.

The overall highest $F_1$-score (0.8654) is achieved by the ensemble classifier MVE in the MSD dataset. While MVE outperforms the individual classifiers in MSMM and MSD datasets, the RF classifier performs best in the MSQ dataset. Notably, in the MSQ dataset, the second-best individual classifier, DT has a significantly low $F_1$-score compared to RF, which potentially leads to the down performance of the ensemble classifier MVE. On the other hand, when both RF and DT have comparable $F_1$-scores, the ensemble classifier can enhance the aggregated performance.

Consistently, the RF, MVE, DT, and LR classifiers secured the top four positions in mean accuracy across all three test datasets, with RF attaining the highest accuracy followed by MVE, DT, and LR. It is noteworthy that while LR and SVC demonstrate reasonable mean accuracy (>0.8), their $F_1$-scores are significantly poor. These observations underscore the potentially misleading nature of the accuracy metric in a classification problem characterized by class imbalance.

It is crucial to highlight that the highest mean values for both $F_1$-score and accuracy are achieved under the MSD dataset. Additionally, across all classifiers, the most optimal classification results are observed in the MSD dataset. Remarkably, the SVC classifier intermittently secured the third rank during $k$-fold cross-validation within the MSD dataset, yet it consistently failed to rank among the top three classifiers in the remaining datasets (see Fig. 5). These findings collectively suggest that the MSD dataset, characterized by more granular and local features, significantly enhances the classification performance of the classifiers. Therefore, superior feature extraction is equally important as superior model selection for optimal results.

Table 2. Classification results of MVE and individual ML classifiers on the 30% testing data from MSMM, MSQ, and MSD datasets, over 100 iterations (*with* cost-sensitive learning).

| Dataset | Classifier | $F_1$-score | | Accuracy | |
|---|---|---|---|---|---|
| | | Mean | Std. dev. | Mean | Std. dev. |
| MSMM | MVE | **0.7725** | 0.0627 | 0.9631 | 0.0106 |
| | RF | 0.7600 | 0.0445 | **0.9710** | 0.0058 |
| | DT | 0.7682 | 0.0646 | 0.9625 | 0.0107 |
| | LR | 0.5333 | 0.0000 | 0.8772 | 0.0000 |
| MSQ | MVE | 0.7733 | 0.0433 | 0.9631 | 0.0070 |
| | RF | **0.8375** | 0.0334 | **0.9799** | 0.0042 |
| | DT | 0.7520 | 0.0611 | 0.9604 | 0.0090 |
| | LR | 0.5333 | 0.0000 | 0.8772 | 0.0000 |
| MSD | MVE | <u>**0.8654**</u> | 0.0629 | 0.9778 | 0.0181 |
| | RF | 0.8552 | 0.0339 | <u>**0.9818**</u> | 0.0045 |
| | DT | 0.8477 | 0.0717 | 0.9759 | 0.0121 |
| | LR | 0.5333 | 0.0000 | 0.8772 | 0.0000 |
| | SVC | 0.4000 | 0.0000 | 0.8421 | 0.0000 |

Table 3. Classification results of MVE and individual ML classifiers on the 30% testing data from MSMM, MSQ, and MSD datasets, over 100 iterations (*without* cost-sensitive learning).

| Dataset | Classifier | $F_1$-score | | Accuracy | |
|---|---|---|---|---|---|
| | | Mean | Std. dev. | Mean | Std. dev. |
| MSMM | MVE | **0.8490** | 0.0238 | **0.9815** | 0.0028 |
| | RF | 0.8339 | 0.0344 | 0.9792 | 0.0046 |
| | DT | 0.6921 | 0.0579 | 0.944 | 0.0111 |
| | LR | 0.7500 | 0.0000 | 0.9649 | 0.0000 |
| MSQ | MVE | 0.8456 | 0.0403 | 0.9804 | 0.0056 |
| | RF | <u>**0.8549**</u> | 0.0113 | <u>**0.9821**</u> | 0.0017 |
| | DT | 0.6586 | 0.0612 | 0.9387 | 0.0120 |
| | LR | 0.8235 | 0.0000 | 0.9737 | 0.0000 |
| MSD | MVE | **0.8264** | 0.0348 | **0.9766** | 0.0041 |
| | RF | 0.7928 | 0.0444 | 0.971 | 0.0062 |
| | DT | 0.8172 | 0.0273 | 0.9737 | 0.0041 |
| | LR | 0.8235 | 0.0000 | 0.9737 | 0.0000 |
| | SVC | 0.7143 | 0.0000 | 0.9649 | 0.0000 |

### 4.2. Machine learning without cost-sensitive learning

To investigate the impact of cost-sensitive learning on classification performance, we replicated the experiments outlined in Section 4.1, excluding the cost-sensitive learning component. The results are reported in Table 3 where the maximum mean values for the metrics within each dataset are emphasized using bold font, and the overall highest mean metric value is underlined. The results in Table 3 reveal crucial insights on the machine learning models and input features.

Cost-sensitive learning notably enhanced the classification performance of DT under all three datasets. This is expected because the algorithm of DT doesn't perform well on imbalanced datasets [25]. While Majority MVE and RF exhibited improved performance on the MSD dataset with cost-sensitive learning, the reverse trend was observed for the MSMM and MSQ datasets. Both LR and SVC demonstrated notably better performance without cost-sensitive learning across all datasets. These findings suggest that the efficacy of cost-sensitive learning varies among different machine learning models. Further exploration is crucial to understand how cost-sensitive learning impacts various machine learning models across multiple imbalanced datasets, presenting an important avenue for future research.

In alignment with the observed trend in cost-sensitive learning, the MVE achieves the highest $F_1$-score compared to individual classifiers on the MSMM and MSD datasets. RF achieves highest $F_1$-score on MSQ dataset, and this is the overall highest score under the setting of without cost-sensitive learning. However, this highest $F_1$-score (0.8549) remains below than the $F_1$-score (0.8654) obtained by the combination of the MSD features, MVE classifier, and cost-sensitive learning.

It is noteworthy that, with the application of cost-sensitive learning, all the ML classifiers attain their highest $F_1$-scores on the MSD dataset (see Table 2). While it might appear that the greater number of feature samples in MSD contributed to this enhanced performance, the absence of cost-sensitive learning reveals a different outcome: RF and MVE achieve their highest $F_1$-scores on the MSQ and MSMM datasets, respectively (see Table 3). These observations underscore that simply increasing the number of features does not guarantee improved model performance. Instead, the effectiveness of input features is intricately linked to the holistic machine learning framework.

*4.3. Benchmarking against existing study*

To validate the effectiveness of our proposed machine learning methodology, MVE, and feature extraction approach, MSD, we conducted a benchmarking analysis against the methodology presented by Chen et al. (2023) [17]. In alignment with their approach, we replicated several key elements: exclusion of unexposed block layers from the classification problem, mitigation of class imbalance through undersampling (utilizing 27 instances of the "Anomalous" class and randomly selecting 27 (out of 297) instances from the bulk "Nominal" class), a 70%-30% split for training and testing data, adoption of 5-fold cross-validation, and the execution of 100 iterations of the entire procedure. The findings from our analysis, alongside the results obtained by Chen et al. (2023) [17], are presented in Table 4.

Table 4 showcases the superior performance of our methodology, demonstrating a 9.66% enhancement in the $F_1$-score compared to the methodology by Chen et al. (2023) [17]. Notably, our MVE-based framework exhibited a lower mean $F_1$-score with a higher standard deviation when employing undersampling techniques, in contrast to the utilization of cost-sensitive learning (refer to Table 2). This discrepancy underscores the effectiveness of cost-sensitive learning as a superior method for addressing the class imbalance, a conclusion supported by several additional empirical studies [26, 27].

Table 4. Comparison of the proposed MVE technique and MSD feature extraction with Chen et al.'s methodology [17].

| Methodology | $F_1$-score | | Accuracy | |
|---|---|---|---|---|
| | Mean | Std. dev. | Mean | Std. dev. |
| *Classifier:* Hyperdimensional computing (HDC) framework [17]. *Feature:* Median and std. dev. from the temporal sampling of raw signals with window size 20. | 0.745 | 0.214 | 0.805 | 0.163 |
| *Classifier:* Proposed MVE-based framework. *Feature:* Mean, std. dev., and deciles (MSD) from layer-wise signals. | 0.817 | 0.096 | 0.824 | 0.080 |

*4.4. Guidelines for practitioners*

The findings in our study provide several practical insights for practitioners, which are summarized as follows:

- While ensemble learning has the potential to outperform individual classifiers, its efficacy depends on the quality of the constituent classifiers. Blindly employing ensemble learning may not yield superior results if only one classifier excels while the rest underperform.
- In the domain of anomaly detection, the use of accuracy as an evaluation metric can be misleading due to the presence of class imbalance.
- Meaningful feature extraction is equally important as effective model selection for optimal results. Both elements contribute synergistically to the overall effectiveness of the machine learning process.
- The effectiveness of input features is intricately linked to the holistic machine learning framework. For instance, in our investigation, the optimal performance was attained using the MSD dataset with the MVE classifier and cost-sensitive learning. However, when cost-sensitive learning was excluded, the RF classifier achieved superior performance with the MSQ dataset. Therefore, it is important to investigate various settings of machine learning frameworks for optimal results.
- Mitigating class imbalance using cost-sensitive learning can yield diverse effects on different machine learning models. In our investigation, while it enhanced the performance of the decision tree classifier, it led to a degradation in the performance of the logistic regression and support vector classifiers. Thus, it is advisable to evaluate machine learning models both with and without cost-sensitive learning to determine the most effective approach.

## 5. Conclusion

Overheating detection is essential for the LPBF process, as it enables corrective actions through process control to ensure high-quality parts. In this paper, we propose a robust machine learning-based framework for detecting overheating anomalies in LPBF at the layer level. We use a benchmark dataset of photodiode measurements from a specimen with intentional overheating anomalies. We extract three sets of features from the raw photodiode data: (mean, standard deviation, median, maximum), MSQ (mean, standard deviation, quartiles), and MSD (mean, standard deviation, deciles). We train cost-sensitive machine learning (ML) classifiers on these features and combine them into a majority voting ensemble (MVE) classifier. Our results demonstrate that the MSD features yield the best performance for all classifiers, and the MVE classifier surpasses the individual ML classifiers. Moreover, our machine learning methodology achieves superior results (9.66% improvement in $F_1$-score) in detecting overheating anomalies, surpassing the other methods in the literature that use the same benchmark dataset. We also offer several valuable insights for LPBF practitioners based on our results.

This research relies on the class labels in the benchmark dataset as the ground truth for the overheating and normal layers. This is a limitation, as the class labels may not be accurate or reliable. Validating the ground truth with other optical and thermal sensors could improve the quality of the class labels and the machine learning models. Hence, data fusion from diverse sensors to create ground truth labels for overheating detection at the layer level is a valuable research direction for developing robust machine learning models.

While our current model shows potential in detecting overheating anomalies across different levels under controlled conditions, further experimentations are imperative to assess its adaptability to real-world LPBF scenarios. In our future research, we plan to evaluate the efficacy of our machine learning framework by collecting data from strategically designed specimens. These specimens will incorporate unexposed blocks with diverse cross-sectional areas, volumes, shapes, and locations, thereby simulating a comprehensive range of overheating anomalies encountered in practical LPBF fabrications.

## Acknowledgements

This research was funded by the National Aeronautics and Space Administration (NASA) under Grant No. 80NSSC21M0100.